\documentclass[a4paper]{article}

\usepackage{INTERSPEECH2020}
\usepackage[font={small,it}]{caption}
\usepackage{multirow}
\usepackage{color}
\newcommand{\checkR}[1]{\textcolor{black}{#1}}

\title{Mel-spectrogram augmentation for sequence-to-sequence voice conversion}
\name{Yeongtae Hwang$^1$, Hyemin Cho$^2$, Hongsun Yang$^3$, Dong-Ok Won$^4$, \\ Insoo Oh$^2$, and Seong-Whan Lee$^{4,5}$}
\address{$^1$Research team, Neosapience, Seoul, Korea \\
$^2$Magellan, Netmarble, Seoul, Korea\\
$^3$AI research, KT Corporation, Seoul, Korea\\
$^4$Department of Brain and Cognitive Engineering, Korea University, Seoul, Korea\\
$^5$Department of Artificial Intelligence, Korea University, Seoul, Korea}
\email{ythwang@neosapience.com, chme@netmarble.com,  hongsun.yang@kt.com, wondongok@korea.ac.kr, ioh@netmarble.com, sw.lee@korea.ac.kr}
\begin{document}

\maketitle 
\begin{abstract}
\checkR{For} training the sequence-to-sequence voice conversion model, we need to handle an issue of insufficient data about the number of speech pairs which consist of the same utterance. 
\checkR{This study experimentally investigated the effects of Mel-spectrogram augmentation on training the sequence-to-sequence voice conversion (VC) model from scratch.} 
For Mel-spectrogram augmentation, we adopted the policies proposed in SpecAugment \cite{park2019specaugment}.
In addition, we proposed new policies \checkR{(i.e., frequency warping, loudness and time length control)} for more data variations . 
\checkR{Moreover,} to find the \checkR{appropriate} hyperparameters of augmentation policies without training the VC model, we \checkR{proposed hyperparameter search strategy and the new metric \checkR{for reducing experimental cost}, namely deformation per deteriorating ratio.} 
We \checkR{compared the effect of these Mel-spectrogram augmentation methods} based on various sizes of training set and augmentation \checkR{policies}.
In the experimental results, the time axis warping based policies \checkR{(i.e., time length control and time warping.)} showed better performance than other policies. 
\checkR{These results indicate that the use of the Mel-spectrogram augmentation is more beneficial for training the VC model.} 

\end{abstract}


\vspace{-0.1cm}
\section{Introduction}
\vspace{-0.1cm}
\label{sec:intro}

Recently developed speech-synthesis techniques \cite{wang2017tacotron, shen2018natural} can produce synthesized speech close to that of the target speaker.
The one of the most reasons for this recent success is that encoder-decoder models with attention mechanisms have been adapted to text-to-speech (TTS) model.
Speaker-adaptation has been investigated to leverage a large amount of speech data to generate a synthesized voice for a new speaker \cite{jia2018transfer, chen2018sample}.
These studies showed impressive results in which synthesized voices are generated by adaptation using a few samples.
\\\indent
Voice conversion (VC) is another speech-synthesis technique.
The purpose of VC is to switch the speech of a source speaker into that of a target without changing the linguistic content.
It acts in a similar manner to the speaker-adaptation technique if it is attached to a TTS system.
In the frame-to-frame VC approaches based on acoustic models, i.e,. joint density Gaussian mixture models (JD-GMM) \cite{kain1998spectral, toda2007voice}, deep neural networks (DNN) \cite{desai2010spectral, chen2014voice} and recurrent neural networks (RNN) \cite{sun2015voice, lai2016phone}, frame alignment using dynamic time warping algorithms must be used during training. 
The application of the encoder-decoder models to VC generates highly intelligible speech without frame alignment.
More recently, a variety of techniques have been proposed to improve sequence-to-sequence(Seq2Seq) VC by adding bottleneck features \cite{zhang2019sequence, zhang2019improving} and text supervision \cite{zhang2019improving, biadsy2019parrotron}.

\begin{figure}[t!]
\begin{minipage}[b]{1.0\linewidth}
\centering
\centerline{\includegraphics[width=7.5cm]{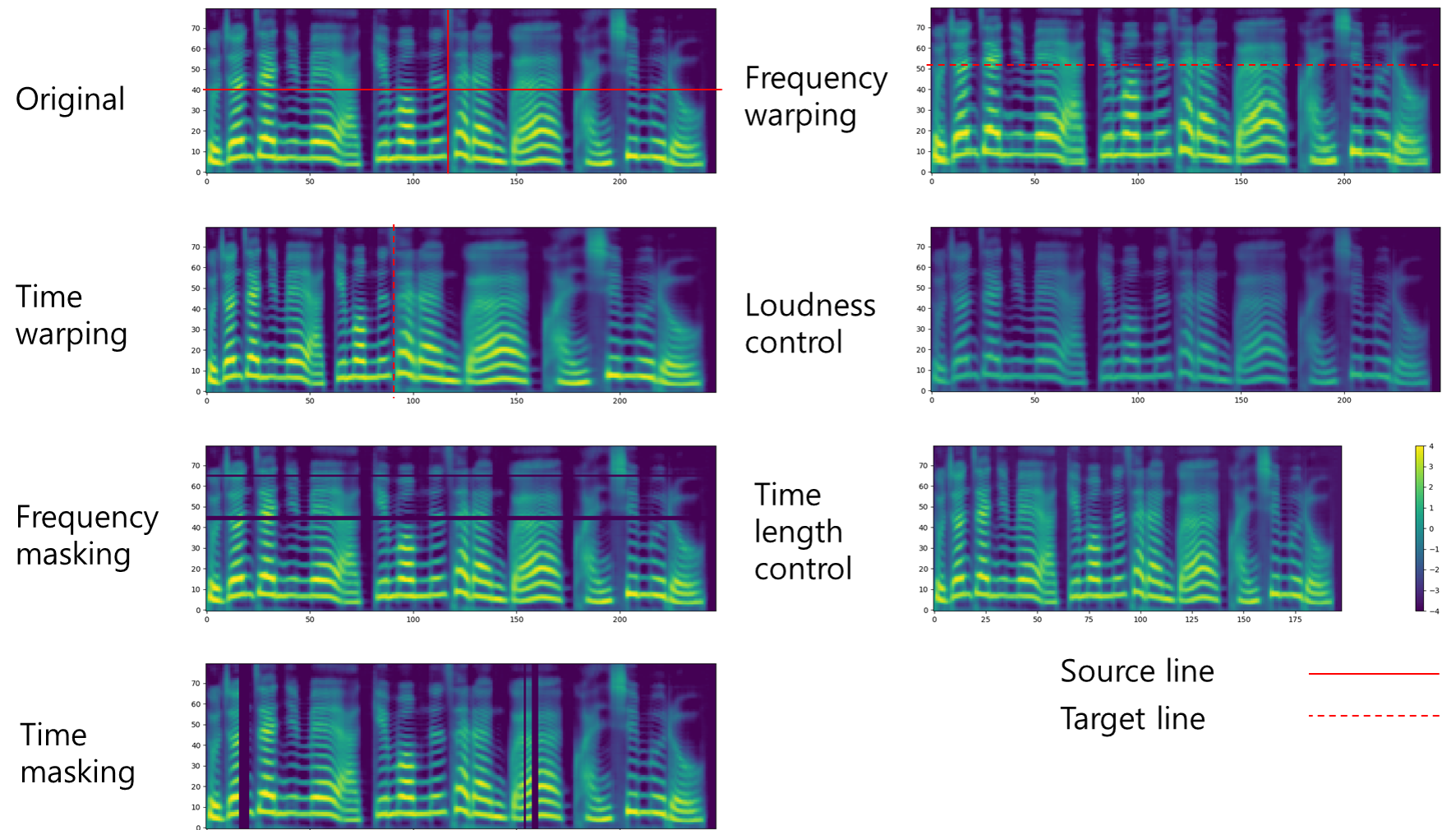}}
\centerline{}\medskip
\end{minipage}

\caption{Example Mel-spectrograms for each augmentation policy. Left: original Mel-spectrogram and transformed Mel-spectrograms with the policies proposed in SpecAugment, namely time warping, frequency masking, and time masking. Right: transformed Mel-spectrograms with the policies proposed herein, namely frequency warping, loudness control, and time length control.}
\label{fig:melaug}
\vspace{-0.4cm}
\end{figure}
Thus far, the main problem with VC is the lack of data consisting of speech pairs containing the same utterance. 
To overcome this situation, data augmentation approaches have been studied based on audio processing \cite{arakawa2019implementation, nachmani2019unsupervised}, text alignment \cite{zhang2019improving}, and synthetic data \cite{biadsy2019parrotron}.
Other speech-related fields, i.e., automatic speech recognition, SpecAugment \cite{park2019specaugment}, vocal track length perturbation (VTLP) \cite{jaitly2013vocal}, and improved vocal track length perturbation (IVTLP) \cite{kim2019improved} have been proposed based on Mel-spectrogram processing for data augmentation. 
\\\indent Inspired by these, we set the goal of this paper as the determination of the effectiveness of Mel-spectrogram augmentation for the Seq2Seq VC model. 
Thus, we adopted policies proposed in SpecAugment for VC. We propose new policies for more Mel-spectrogram variants. Choosing hyperparameters for Mel-spectrogram augmentation has a large impact on the Seq2Seq VC model training.
\checkR{To select appropriate hyperparameters for Mel-spectrogram augmentation without any training the VC model, we proposed hyperparameter search strategy based on our proposed metric, namely deformation per deteriorating (DPD) ratio.}
To evaluate the effectiveness of Mel-spectrogram augmentation, we conducted experiments that is one to one VC task with various sizes of training data and augmentation policies. 
In the experimental results, time warping-based policies showed metrics better than other policies. Among them, our proposed time length control was most effective when it applied to the source and target Mel-spectrogram in the same way. The audio samples of this study are shown on our demo web\footnote{Audio samples: https://chmenet.github.io/demo/}. 
\vspace{-0.1cm}
\section{Mel-spectrogram Augmentation}
\vspace{-0.1cm}
\label{sec:melaug}
We adopted policies proposed in SpecAugment, namely, time masking, frequency masking, and time warping, to deform the time axis, partial loss of time axis, and partial loss of frequency axis. For more variety of Mel-spectrogram variants, we propose the new policies of frequency warping, loudness control, and time-length control to adjust the pitch, loudness, and speed of speech. The frequency warping of VTLP and IVTLP is similar to one of our frequency warping cases in which the source frequency point is fixed in the middle of the frequency. Thus, our frequency warping allows for greater frequency variation. Note that the aforementioned policies are applicable online during training.
Fig. \ref{fig:melaug} shows how each policy transforms the Mel-spectrogram.
\vspace{-0.2cm}
\subsection{Augmentation policy} \label{AugPolicy}
Given a Mel-spectrogram with $\tau$ lengths on time axis and $\nu$ lengths on frequency axis, the following policies can be used.\\
{\bf Time warping (TW): }
The source point in the time axis is chosen from $[\lfloor\tau/4\rfloor, \tau-\lfloor\tau/4\rfloor]$. It is to be warped by a time distance $w \in [-W\tau,W\tau]$, where $W$ is the time warp parameter. The voice speeds of the two parts based on the target point differ.
\\
{\bf Frequency masking (FM):}
$f$ consecutive Mel-frequency channels $[f_0, f_0+f)$ is selected, where $f$ is a discrete random variable $\in$ $[$0, $F$$]$, $F$ is the frequency masking parameter, $f_0$ is chosen from $[0, \nu-f]$. Selected region is replaced by the minimum value. This process is repeated $N_f$ times.\\
{\bf Time masking (TM): }
$t$ consecutive time steps $[t_0, t_0+t)$ is selected, where $t$ is a discrete random variable $\in$ $[0, T]$, $T$ is the time masking parameter, $t_0$ is chosen from $[0, \tau-t]$. Selected region is replaced by the minimum value. This process is repeated $N_t$ times.\\
{\bf Frequency warping (FW): }
The source point in the frequency axis is chosen from $[\lfloor\nu/4\rfloor, \nu-\lfloor\nu/4\rfloor]$. The source points with all time points are to be warped by a frequency distance $h \in [-H,H]$, where $H$ is the frequency warp parameter. It increases or decreases the level of the pitch.\\
{\bf Loudness control (LC): }
Subtract the minimum to all Mel-spectro-gram values and multiply them by $1-\lambda$ where $\lambda \in [0,\Lambda]$, $\Lambda$ is the loudness control parameter. Then add the minimum value to them. It makes the loudness of the speech either down or not.\\
{\bf Time length control (TLC): }
The source point in the time axis is $\tau$. A line parallel to the frequency axis with the source point warped by a time distance $l \in [-L\tau,L\tau]$, where $L$ is the time length control parameter. It increases or decreases the speed of the speech.

\vspace{-0.2cm}
\subsection{Deformation per deteriorating ratio}
A good parameter for the Mel-spectrogram augmentation gives maximum variation without losing speech quality. To fit this definition, we propose a new metric, the DPD ratio, which is described by the following equation:
\begin{equation}
	\text{DPD$_p$} = D_p/|E_p - E_o|
	\label{matric}
\end{equation}
where $D_p$ is the maximum ratio of deformation for $p$, $E_p$ is the expectation value of character error rate (CER) for $p$, $P$ is $\{\{T,N_t\}, \{F,N_f\}, W, H, L, \Lambda\}$ the set of hyperparameter for Mel-spectrogram augmentation, $p$ is an element of $P$, $E_o$ is the expectation value of CER without augmentation policy. $|E_p - E_o|$ represents deteriorating effects for each hyperparameter. Table \ref{tab:defp} shows the definition of $D_p$ for each policy.

\begin{table}[t]
\caption{Definition of the maximum ratio of deformation $D_p$, where $p$ is the hyperparameter for the augmentation policy.}
\centering
\setlength\tabcolsep{9.8pt}
\label{tab:defp}
\begin{tabular}{|l|l|l|l|l|l|l|}
\hline
\multicolumn{1}{|c|}{\textbf{$p$}} & \multicolumn{1}{c|}{\textbf{${T,N_t}$}} & \multicolumn{1}{c|}{\textbf{${F,N_f}$}} & \multicolumn{1}{c|}{\textbf{$W$}} & \multicolumn{1}{c|}{\textbf{$H$}} & \multicolumn{1}{c|}{\textbf{$L$}} & \multicolumn{1}{c|}{\textbf{$\Lambda$}} \\ \hline
$D_p$& $\frac{T\times N_t}{E(\tau)}$ & $\frac{F\times N_f}{\nu}$ & $W$ & $\frac{H}{\nu}$ & $L$ & $\Lambda$ \\ \hline
\end{tabular}

\end{table}

\begin{table}[t!]
\caption{DPDs on validation set of KSS dataset. The variables not recorded in the table are as follows. $E_o = 0.201, E(\tau) = 217.0$ and $\nu = 80$. The maximum DPD and the selected hyperparameter for each policy are highlighted in bold. }

\setlength\tabcolsep{1.5pt}

\label{tab:dpd1}
\begin{tabular}{lcccccccc}
\hline
\multicolumn{9}{c}{\textbf{Time masking ($N_t=1$)}} \\ \hline
$T$ & 2 & 4 & 6 & \textbf{8} & 10 & 12 & 14 & 16 \\
$D_{T,N_t}$ & 0.009 & 0.018 & 0.028 & 0.037 & 0.046 & 0.055 & 0.065 & 0.074 \\
$E_{T,N_t}$ & 0.215 & 0.217 & 0.225 & 0.222 & 0.232 & 0.234 & 0.240 & 0.248 \\
DPD$_{T,N_t}$ & 0.643 & 1.125 & 1.167 & \textbf{1.762} & 1.484 & 1.667 & 1.667 & 1.574 \\ \hline
\multicolumn{9}{c}{\textbf{Frequency masking ($N_f=1$)}} \\ \hline
$F$ & 2 & 4 & \textbf{6} & 8 & 10 & 12 & 14 & 16 \\
$D_{F,N_f}$ & 0.025 & 0.050 & 0.075 & 0.100 & 0.125 & 0.150 & 0.175 & 0.200 \\
$E_{F,N_f}$ & 0.217 & 0.227 & 0.235 & 0.271 & 0.266 & 0.302 & 0.340 & 0.347 \\
DPD$_{F,N_f}$ & 1.563 & 1.923 & \textbf{2.206} & 1.429 & 1.923 & 1.485 & 1.259 & 1.370 \\ \hline
\multicolumn{9}{c}{\textbf{Time warping}} \\ \hline
$W$ & 0.020 & 0.040 & 0.060 & \textbf{0.080} & 0.100 & 0.120 & 0.140 & 0.160 \\
$D_{W}$ & 0.020 & 0.040 & 0.060 & 0.080 & 0.100 & 0.120 & 0.140 & 0.160 \\
$E_W$ & 0.218 & 0.217 & 0.220 & 0.223 & 0.242 & 0.256 & 0.265 & 0.280 \\
DPD$_{W}$ & 1.176 & 2.500 & 3.158 & \textbf{3.636} & 2.439 & 2.182 & 2.188 & 2.025 \\ \hline
\multicolumn{9}{c}{\textbf{Frequency warping}} \\ \hline
$H$ & 2 & \textbf{4} & 6 & 8 & 10 & 12 & 14 & 16 \\
$D_H$ & 0.025 & 0.050 & 0.075 & 0.1 & 0.125 & 0.15 & 0.175 & 0.2 \\
$E_H$ & 0.225 & 0.237 & 0.286 & 0.341 & 0.400 & 0.437 & 0.515 & 0.545 \\
DPD$_{H}$ & 1.042 & \textbf{1.389} & 0.882 & 0.714 & 0.628 & 0.636 & 0.557 & 0.581 \\ \hline
\multicolumn{9}{c}{\textbf{Time length control}} \\ \hline
$L$ & 0.020 & 0.040 & 0.060 & 0.080 & 0.100 & \textbf{0.120} & 0.140 & 0.160 \\
$D_L$ & 0.020 & 0.040 & 0.060 & 0.080 & 0.100 & 0.120 & 0.140 & 0.160 \\
$E_L$ & 0.211 & 0.210 & 0.220 & 0.211 & 0.216 & 0.205 & 0.219 & 0.213 \\
DPD$_{L}$ & 2.000 & 4.444 & 3.158 & 8.000 & 6.667 & \textbf{30.00} & 7.778 & 13.333 \\ \hline
\multicolumn{9}{c}{\textbf{Loudness control}} \\ \hline
$\lambda$ & 0.020 & 0.040 & 0.080 & \textbf{0.160} & 0.320 & 0.640 & - & - \\
$D_\lambda$ & 0.020 & 0.040 & 0.080 & 0.160 & 0.320 & 0.640 & - & - \\
$E_\lambda$ & 0.213 & 0.217 & 0.218 & 0.221 & 0.254 & 0.406 & - & - \\
DPD$_{\lambda}$ & 1.667 & 2.500 & 4.706 & \textbf{8.000} & 6.038 & 3.122 & - & - \\ \hline
\end{tabular}

\end{table}

\begin{table}[t!]
\caption{DPDs on validation set of KSS dataset for masking policies. The maximum DPD and the selected hyperparameter for each policy are highlighted in bold. }

\label{tab:dpd2}
\setlength\tabcolsep{11.0pt}
\begin{tabular}{lcccc}
\hline
\multicolumn{5}{c}{\textbf{Time masking}} \\ \hline
$T,N_t$ & 1,8 & 2,4 & \textbf{4,2} & 8,1 \\
$D_{T,N_t}$ & 0.037 & 0.037 & 0.037 & 0.037 \\
$E_{T,N_t}$ & 0.216 & 0.218 & 0.212 & 0.222 \\
DPD$_{T,N_t}$ & 2.467 & 2.176 & \textbf{3.364} & 1.762 \\ \hline
\multicolumn{5}{c}{\textbf{Frequency masking}} \\ \hline
$F,N_f$ & 1,6 & 2,3 & \textbf{3,2} & 6,1 \\
$D_{F,N_f}$ & 0.075 & 0.075 & 0.075 & 0.075 \\
$E_{F,N_f}$ & 0.218 & 0.213 & 0.212 & 0.235 \\
DPD$_{F,N_f}$ & 4.412 & 6.250 & \textbf{6.818} & 2.206 \\ \hline
\end{tabular}
\vspace{-0.4cm}
\end{table}
\vspace{-0.2cm}
\subsection{Hyperparameter search \checkR{based on DPD}}
\checkR{In general, optimal parameters can be derived from experimental results with training the model. But this is expensive to perform if there are many types of parameters or many types of experiments. We propose a new hyperparameter search strategy based on the DPD ratio to find appropriate hyperparameter for the Mel-spectrogram augmentation without any training the VC model.}\\
\indent The voices for searching the \checkR{appropriate} hyperparameter were 64 audio of the Korean single speaker speech (KSS) datasets \cite{KSSdataset}. These selected audio were converted to Mel-spectrograms.
The Mel-spectrogram augmentation for each hyperparameter was performed 10 times to compute $E_p$.
Because Korean is sensitive to spacing, CER is more reliable for $E_p$ than word error rate (WER).
CER was calculated using the recognition result of the Google Speech API. 
The audio for computing CER was decoded using Griffin-Lim \cite{griffin1984signal} vocoder from a Mel-spectrogram with or without doing augmentation. Through $E_p$ in Table \ref{tab:dpd1}, you can see the degree of deterioration by adjusting $p$ for each policy. \\  
\indent In this experiment, p was increased in the arithmetic sequence except $\Lambda$.
Because the policy LC only controls audio volume and there is substantial difference in CER performance according to adjust $\Lambda$.
Thus, $\Lambda$ was increased to a geometric sequence in this experiment.
The hyperparameters determined by choosing the maximum {DPD$_p$} are shown in Table \ref{tab:dpd1}.
In addition, Time masking and frequency masking have two hyperparameters.
To determine the \checkR{appropriate} combination for these, we first set $N_f$ and $N_t$ to one to find the best $D_{T,N_t}$ and $D_{F,N_f}$.
With fixed $D_{T,N_t}$ and $D_{F,N_f}$ values, we experimented with all possible combinations for $T,N_t$ and $F,N_f$. Table \ref{tab:dpd2} shows the best $T,N_t$ and $F,N_f$ to maximize {DPD$_{T,N_t}$} and {DPD$_{F,N_f}$}. 
The determined hyperparameters are in bold in Tables \ref{tab:dpd1} and \ref{tab:dpd2}. In addition, all of them are used in further experiments to evaluate the efficiency of the Mel-spectrogram augmentation.
\vspace{-0.1cm}
\section{Voice Conversion Model}
\vspace{-0.1cm}
\label{sec:model}
We used a simple model, independent of other models to extract bottleneck features and phoneme labels. Our VC model is based on Tacotron2. The input and output of this model are the Mel-spectrogram. The layers in attention and decoder are the same architecture of Tacotron2. Only the encoder has been modified to FC, FC, and LSTM in a similar manner to the decoder. This is because the prenet effectively represents the Mel-spectrograms.
The number of nodes in the layer was determined by referring to the SCENT \cite{zhang2019sequence}. The model configurations are shown in Table \ref{tab:config}. The final waveform is generated using the Wavenet \cite{oord2016wavenet} neural vocoder conditioned on the Mel-spectrogram.  
\vspace{-0.1cm}
\section{Experimental Result}
\vspace{-0.1cm}
\subsection{Experimental condition}

Two datasets were used in our experiment. For the source speaker, we used the KSS dataset, which consists of 12,853 Korean utterances from a female speaker (approximately 12+ hours). For the target speaker, we used an internal dataset by recording based on the transcript of the KSS dataset from a female speaker. After trimming the silence of both, the pair dataset is constructed with containing 12,798 utterances (approximately 8+ hours for each). We used 64 utterances as the validation set and 64 utterances as the test set; the rest were used as training sets. \\\indent
All of VC networks were trained for $1\times10^5$ iterations using the Adam optimizer \cite{kingma2014adam}, with a batch size of 32 and a step size of $1\times10^{-3}$. Wavenet networks were trained for for $16\times10^4$ iterations using the Adam optimizer, with 8bit mu-law quantization for audio amplitude, a batch size of 16 and a step size of $1\times10^{-3}$. 

\begin{table}[t]
\caption{Details of model configurations.}

\setlength\tabcolsep{1.7pt}
\label{tab:config}
\begin{tabular}{|l|l|l|}
\hline
\multirow{4}{*}{VC} & Encoder & \begin{tabular}[c]{@{}l@{}}FC-ReLU-Dropout(0.5), 256 cells $\times 2$\\ Forward-LSTM, 256 cells\end{tabular} \\ \cline{2-3}
& PreNet & FC-ReLU-Dropout(0.5), 256 cells $\times 2$ \\ \cline{2-3}
& Decoder & \begin{tabular}[c]{@{}l@{}}Attention LSTM, 256 cells;\\ Decoder LSTM, 256 cells;\\ Linear project FC, 80 cells;\\ Gate FC, 1 cell and sigmoid activation\end{tabular} \\ \cline{2-3}
& PostNet & \begin{tabular}[c]{@{}l@{}}1D convolution-BN-ReLU-Dropout(0.5),\\ $\quad$256 channels and 5 kernels $\times 4$;\\ 1D convolution-BN-ReLU-Dropout(0.5),\\ $\quad$80 channels and 5 kernels\end{tabular} \\ \hline
\multirow{2}{*}{Vocoder} & Upsampling & \begin{tabular}[c]{@{}l@{}}Subpixel \cite{shi2016deconvolution} convolution, \\ $\quad$3$\times$3 kernels and 1$\times$11 strides;\\ Subpixel convolution, \\ $\quad$3$\times$3 kernels and 1$\times$25 strides\end{tabular} \\ \cline{2-3}
& WaveNet & \begin{tabular}[c]{@{}l@{}}20 layers dilated convolution layers,\\ $\quad$with dilation $d=2^{k \,\text{mod}\,10}$ for \\ $\quad$$k=[0, ... ,19]$, 256 softmax output\end{tabular} \\ \hline
\end{tabular}
\textit{FC represents fully connected, LSTM represents long short-term memory, BN represents batch normalization, ReLU represents rectified linear unit. }
\vspace{-0.4cm}
\end{table}

\vspace{-0.2cm}
\subsection{Evaluation metric}
\checkR{
\textbf{Objective evaluation:} Researches \cite{zhang2019sequence, zhang2019improving} have been adopted mel-cepstrum distortion (MCD) as an evaluation metric to evaluate the acoustic \checkR{similarity} between the synthesized audio and the target audio.
To measure the linguistic expressiveness, one VC study \cite{keskin2019measuring} used ARS metrics, such as WER and CER.
With reference to the aforementioned studies, MCD and CER were adopted as metrics for objective evaluation. \\
\noindent \textbf{Subjective evaluation:}  Two subjective evaluations were conducted to evaluate the perceptual performance by the mean opinion score (MOS) test on the Likert scale, participated in 20 Korean native subjects. 
The first evaluation is \checkR{naturalness}. For this task, participants were asked to evaluate the \checkR{naturalness} of the generated speech.
The second evaluation is the \checkR{similarity}, which to evaluates how similar the generated voice to the target speaker. \\
\indent MCD, CER, \checkR{naturalness} and \checkR{similarity} were reported on the test set in Tables \ref{tab:evalvol}, \ref{tab:singleaug}, 
and  \ref{tab:policy-volumes}.
}
\vspace{-0.2cm}
\subsection{Baseline performance}
In order to observe the performance change of VC model according to data usage without Mel-spectrogram augmentation, we experimented by reducing the number of training data to half of it each time from the whole training set till it reaches to the 1/16 training set. When using lower number of training data than 1/16, the VC model shows unstable that often failed to make attention values close to diagonal.
\\\indent
Table \ref{tab:evalvol} shows the results. 
We set them to the baseline performance for each size of training set. The metrics obtained in this experiment were used as a criterion for determining the degree to which the augmentation policy has improved performance. 
The CER performance is directly proportional to the amount of training data.
However, MCD, \checkR{naturalness} and \checkR{similarity} are not directly related to the amount of training data.
\vspace{-0.2cm}
\subsection{Effectiveness of augmentation policy}
In this experiment, all augmentation policies were applied to the source Mel-spectrogram. One-to-many mapping data in the training set makes the model difficult to converge. In general, the augmentation is not applied to target data. However, if the speeds of the source audio and target audio are changed to the same ratio, this is one-to-one mapping and means augmenting pair data. Therefore, we experimented with two cases, namely, applying TLC only to source audio, and applying TLC to both source and target. The second case is denoted `TLC both.'
\begin{table}[t]
\caption{Evaluation results using various sizes of training data without Mel-spectrogram augmentation, in addition natural speech of source and target speaker. We set them to the baseline performance for each size of training set.} 
\label{tab:evalvol}

\setlength\tabcolsep{2.2pt}
\begin{tabular}{llllllll}
\hline
  & Source & Target & 1 & 1/2 & 1/4 & 1/8 & 1/16 \\ \hline
\multirow{4}{*}{\begin{tabular}[c]{l}
\end{tabular}}MCD & - & - & 6.873 & 7.123 & 6.759 & 6.850 & 7.367 \\
CER & 0.066 & 0.145 & 0.143 & 0.159 & 0.225 & 0.323 & 0.479 \\
\checkR{Naturalness} & 4.528 & 4.493 & 3.363 & 3.448 & 3.416 & 3.229 & 2.460 \\
\checkR{\checkR{Similarity}} & 2.202 & 4.192 & 3.550 & 3.466 & 3.594 & 3.470 & 2.892 \\ \hline
\end{tabular}

\end{table}

\begin{table}[t!]
\caption{Evaluation results by applying a single augmentation policy on the 1/16 training set. The better metrics against baseline performance on the 1/16 training set are highlighted in bold.}
\vspace{-0.4cm}
\label{tab:singleaug}
\setlength\tabcolsep{1.3pt}

\begin{tabular}{llllllllll}
\hline
  & 1/16 &TW& FM & TM & FW & LC & TLC & \begin{tabular}[c]{@{}l@{}}TLC\\ both\end{tabular} \\ \hline
\multirow{3}{*}{\begin{tabular}[c]{l}\end{tabular}}{MCD} & {7.367} & \textbf{7.281} & {7.439} & {7.512} & {7.572} & {7.401} & {\textbf{7.318}} & {{7.392}} \\
{CER} & {0.479} & \textbf{0.423} & {0.547} & {0.575} & {0.641} & {0.489} & \textbf{0.426} & {\textbf{0.397}} \\
{\checkR{Naturalness}} & {2.460} & \textbf{2.481} & {2.402} & {2.191} & {2.271} & {2.439} & {\textbf{2.523}} & {\textbf{2.545}} \\
{\checkR{Similarity}} & {2.892} & \textbf{2.914} & {2.869} & {2.712} & {2.752} & \textbf{2.893} & {\textbf{2.938}} & {\textbf{2.964}} \\\hline
\end{tabular}
\vspace{-0.3cm}
\end{table}
\noindent{\bf Single policy:} To verify the effectiveness of each augmentation policy, we experimented with the 1/16 training set.
The results for each policy are shown in Table \ref{tab:singleaug}.
Policies showing improved metrics against baseline performance on the 1/16 training set were TLC, `TLC both,' and TW. 
Policies based on the time axis warping cause differences in the speed of speech.
Improved performances can be interpreted as policies based on the time axis can generate different distributions to source speech with less loss of speech characteristics of the speaker.
Masking policies hinder learning because it gives a loss of information in the source.
Frequency axis warping produces a phonetic distribution that differs from the actual speaker, which seems to adversely affect the conversion using the actual speaker's speech. 
LC only reduces the Mel-spectrogram value. Thus, it shows a similar performance that of the baseline.\\
{\bf Multiple policy:} 
\checkR{We conducted an experiment combining TLC, `TLC both,' TW, and LC that improved or preserved performance on the single policy experiment. 
However, we observed that these combinations could not improved performance.
We assume the reason for this results that using multiple policies for Mel-spectrogram augmentation led to the loss of linguistic and acoustic information of the original speaker. 
In future work, there is a need to check if there is any improvement in performance when we choose appropriate hyperparameter for multiple policies based on DPD.}\\
{\bf Policy effectiveness:} 
TW, TLC and `TLC both' were tested on all sizes of training data in Table \ref{tab:evalvol}. 
The results are shown in Table \ref{tab:evalvol}. 
\checkR{
Those show slight improvement or similar performance without depending on the number of training data. 
}
\checkR{
In experiments for all sizes of training set, the best metrics were mostly in `TLC both.' We guess the reason for this as follows. `TLC both' is able to generate one to one data from the original speech pair. It effectively works than other policies.
}
Therefore, applying `TLC both' to Seq2Seq VC could be an good candidate to improve the performance.

\begin{table}[t!]
\caption{Evaluation results by applying each augmentation policy on various data volumes. The better metrics against baseline performance on each volume dataset are highlighted in bold. The best metrics within results on the same size of training set are highlighted in underlines.}

\label{tab:policy-volumes}
\setlength\tabcolsep{4.3pt}

\begin{tabular}{lllllll}
\hline
Policy & Size & 1 & 1/2 & 1/4 & 1/8 & 1/16 \\ \hline
\multirow{4}{*}{\begin{tabular}[c]{l} - \end{tabular}}
& MCD & 6.873 & 7.123 & 6.759 & 6.850 & 7.367 \\
& CER & 0.143 & 0.159 & 0.225 & 0.323 & 0.479 \\
& \checkR{Naturalness} & 3.363 & 3.448 & 3.416 & 3.229 & 2.460 \\
& \checkR{Similarity} & 3.550 & 3.466 & 3.594 & 3.470 & 2.892 \\ \hline
\multirow{4}{*}{\begin{tabular}[c]{@{}l@{}}TW \end{tabular}}
& MCD & \textbf{6.852} & \underline{\textbf{6.935}} & \textbf{6.692} & \textbf{6.820} & \underline{\textbf{7.281}} \\
& CER & 0.158 & \underline{\textbf{0.143}} & 0.236 & \textbf{0.308} & \textbf{0.423} \\
& \checkR{Naturalness} & \textbf{3.391} & \underline{\textbf{3.516}} & 3.393 & \textbf{3.248} & \textbf{2.481} \\
& \checkR{Similarity} & 3.535 & \underline{\textbf{3.598}} & \textbf{3.597} & \textbf{3.484} & \textbf{2.914} \\ \hline
\multirow{4}{*}{\begin{tabular}[c]{@{}l@{}}TLC \end{tabular}}
& MCD & \textbf{6.829} & 7.126 & \underline{\textbf{6.680}} & 6.997 & \textbf{7.318} \\
& CER & 0.167 & \textbf{0.158} & \textbf{0.202} & \textbf{0.290} & \textbf{0.426} \\
& \checkR{Naturalness} & \textbf{3.372} & 3.421 & \textbf{3.498} & \textbf{3.234} & \textbf{2.523} \\
& \checkR{Similarity} & \textbf{3.577} & \textbf{3.507} & \underline{\textbf{3.609}} & \underline{\textbf{3.491}} & \textbf{2.938} \\\hline
\multirow{4}{*}{\begin{tabular}[c]{@{}l@{}}TLC\\ both \end{tabular}}
& MCD & \underline{\textbf{6.641}} & \textbf{7.052} & \textbf{6.690} & \underline{\textbf{6.740}} & 7.392 \\
& CER & \underline{\textbf{0.134}} & 0.198 & \underline{\textbf{0.185}} & \underline{\textbf{0.282}} & \underline{\textbf{0.397}} \\
& \checkR{Naturalness} & \underline{\textbf{3.540}} & 3.281 & \underline{\textbf{3.509}} & \underline{\textbf{3.253}} & \underline{\textbf{2.545}} \\
& \checkR{Similarity} & \underline{\textbf{3.673}} & 3.400 & \textbf{3.598} & 3.468 & \underline{\textbf{2.964}} \\\hline
\end{tabular}
\vspace{-0.4cm}
\end{table}
\vspace{-0.2cm}
\section{Conclusion}
\vspace{-0.1cm}
This paper describes the effect of Mel-spectrogram augmentation on the one-to-one Seq2Seq VC model. 
\checkR{We adopted policies from SpecAugment and proposed new policies for Mel-spectrogram augmentation. We selected appropriate hyperparameters for each policy through experiments based on our proposed DPD metric without training the VC model. 
Our VC experiments examined the relationship between the size of training sets, the characteristics of augmentation policies, and the resulting VC quality. 
In addition, the time axis warping based policies showed better performance than other policies. 
These results indicate that the use of policies based on the time axis warping is more efficiently training for developing the VC model.}

\section{Acknowledgement}
This work was supported by Netmarble, and Institute of Information $\&$ communications Technology Planning $\&$ Evaluation (IITP) grant funded by the Korea government (MSIT) (No. 2019-0-00079. Department of Artificial Intelligence, Korea University).
We would like to cordially thank Shounan An, Youshin Lim, Seulji Lee, and Kwangyong Lee for their valuable comments on the manuscript. 

\bibliographystyle{IEEEtran}

\bibliography{main}

\begin{thebibliography}{10}
\providecommand{\url}[1]{#1}
\csname url@samestyle\endcsname
\providecommand{\newblock}{\relax}
\providecommand{\bibinfo}[2]{#2}
\providecommand{\BIBentrySTDinterwordspacing}{\spaceskip=0pt\relax}
\providecommand{\BIBentryALTinterwordstretchfactor}{4}
\providecommand{\BIBentryALTinterwordspacing}{\spaceskip=\fontdimen2\font plus
\BIBentryALTinterwordstretchfactor\fontdimen3\font minus
  \fontdimen4\font\relax}
\providecommand{\BIBforeignlanguage}[2]{{%
\expandafter\ifx\csname l@#1\endcsname\relax
\typeout{** WARNING: IEEEtran.bst: No hyphenation pattern has been}%
\typeout{** loaded for the language `#1'. Using the pattern for}%
\typeout{** the default language instead.}%
\else
\language=\csname l@#1\endcsname
\fi
#2}}
\providecommand{\BIBdecl}{\relax}
\BIBdecl

\bibitem{park2019specaugment}
D.~S. Park, W.~Chan, Y.~Zhang, C.-C. Chiu, B.~Zoph, E.~D. Cubuk, and Q.~V. Le,
  ``Specaugment: A simple data augmentation method for automatic speech
  recognition,'' \emph{arXiv preprint arXiv:1904.08779}, 2019.

\bibitem{wang2017tacotron}
Y.~Wang, R.~Skerry-Ryan, D.~Stanton, Y.~Wu, R.~J. Weiss, N.~Jaitly, Z.~Yang,
  Y.~Xiao, Z.~Chen, S.~Bengio \emph{et~al.}, ``Tacotron: Towards end-to-end
  speech synthesis,'' \emph{arXiv preprint arXiv:1703.10135}, 2017.

\bibitem{shen2018natural}
J.~Shen, R.~Pang, R.~J. Weiss, M.~Schuster, N.~Jaitly, Z.~Yang, Z.~Chen,
  Y.~Zhang, Y.~Wang, R.~Skerrv-Ryan \emph{et~al.}, ``Natural tts synthesis by
  conditioning wavenet on mel spectrogram predictions,'' in \emph{International
  Conference on Acoustics, Speech and Signal Processing (ICASSP)}.\hskip 1em
  plus 0.5em minus 0.4em\relax IEEE, 2018, pp. 4779--4783.

\bibitem{jia2018transfer}
Y.~Jia, Y.~Zhang, R.~Weiss, Q.~Wang, J.~Shen, F.~Ren, P.~Nguyen, R.~Pang, I.~L.
  Moreno, Y.~Wu \emph{et~al.}, ``Transfer learning from speaker verification to
  multispeaker text-to-speech synthesis,'' in \emph{Advances in neural
  information processing systems}, 2018, pp. 4480--4490.

\bibitem{chen2018sample}
Y.~Chen, Y.~Assael, B.~Shillingford, D.~Budden, S.~Reed, H.~Zen, Q.~Wang, L.~C.
  Cobo, A.~Trask, B.~Laurie \emph{et~al.}, ``Sample efficient adaptive
  text-to-speech,'' \emph{arXiv preprint arXiv:1809.10460}, 2018.

\bibitem{kain1998spectral}
A.~Kain and M.~W. Macon, ``Spectral voice conversion for text-to-speech
  synthesis,'' in \emph{International Conference on Acoustics, Speech and
  Signal Processing (ICASSP)}, vol.~1.\hskip 1em plus 0.5em minus 0.4em\relax
  IEEE, 1998, pp. 285--288.

\bibitem{toda2007voice}
T.~Toda, A.~W. Black, and K.~Tokuda, ``Voice conversion based on
  maximum-likelihood estimation of spectral parameter trajectory,''
  \emph{Transactions on Audio, Speech, and Language Processing}, vol.~15,
  no.~8, pp. 2222--2235, 2007.

\bibitem{desai2010spectral}
S.~Desai, A.~W. Black, B.~Yegnanarayana, and K.~Prahallad, ``Spectral mapping
  using artificial neural networks for voice conversion,'' \emph{Transactions
  on Audio, Speech, and Language Processing}, vol.~18, no.~5, pp. 954--964,
  2010.

\bibitem{chen2014voice}
L.-H. Chen, Z.-H. Ling, L.-J. Liu, and L.-R. Dai, ``Voice conversion using deep
  neural networks with layer-wise generative training,'' \emph{ACM Transactions
  on Audio, Speech and Language Processing (TASLP)}, vol.~22, no.~12, pp.
  1859--1872, 2014.

\bibitem{sun2015voice}
L.~Sun, S.~Kang, K.~Li, and H.~Meng, ``Voice conversion using deep
  bidirectional long short-term memory based recurrent neural networks,'' in
  \emph{International Conference on Acoustics, Speech and Signal Processing
  (ICASSP)}.\hskip 1em plus 0.5em minus 0.4em\relax IEEE, 2015, pp. 4869--4873.

\bibitem{lai2016phone}
J.~Lai, B.~Chen, T.~Tan, S.~Tong, and K.~Yu, ``Phone-aware lstm-rnn for voice
  conversion,'' in \emph{International Conference on Signal Processing
  (ICSP)}.\hskip 1em plus 0.5em minus 0.4em\relax IEEE, 2016, pp. 177--182.

\bibitem{zhang2019sequence}
J.-X. Zhang, Z.-H. Ling, L.-J. Liu, Y.~Jiang, and L.-R. Dai,
  ``Sequence-to-sequence acoustic modeling for voice conversion,'' \emph{ACM
  Transactions on Audio, Speech and Language Processing (TASLP)}, vol.~27,
  no.~3, pp. 631--644, 2019.

\bibitem{zhang2019improving}
J.-X. Zhang, Z.-H. Ling, Y.~Jiang, L.-J. Liu, C.~Liang, and L.-R. Dai,
  ``Improving sequence-to-sequence voice conversion by adding
  text-supervision,'' in \emph{International Conference on Acoustics, Speech
  and Signal Processing (ICASSP)}.\hskip 1em plus 0.5em minus 0.4em\relax IEEE,
  2019, pp. 6785--6789.

\bibitem{biadsy2019parrotron}
F.~Biadsy, R.~J. Weiss, P.~J. Moreno, D.~Kanvesky, and Y.~Jia, ``Parrotron: An
  end-to-end speech-to-speech conversion model and its applications to
  hearing-impaired speech and speech separation,'' \emph{arXiv preprint
  arXiv:1904.04169}, 2019.

\bibitem{arakawa2019implementation}
R.~Arakawa, S.~Takamichi, and H.~Saruwatari, ``Implementation of dnn-based
  real-time voice conversion and its improvements by audio data augmentation
  and mask-shaped device,'' in \emph{The 10th ISCA Speech Synthesis Workshop
  (to appear)}, 2019.

\bibitem{nachmani2019unsupervised}
E.~Nachmani and L.~Wolf, ``Unsupervised singing voice conversion,'' \emph{arXiv
  preprint arXiv:1904.06590}, 2019.

\bibitem{jaitly2013vocal}
N.~Jaitly and G.~E. Hinton, ``Vocal tract length perturbation (vtlp) improves
  speech recognition,'' in \emph{Proc. ICML Workshop on Deep Learning for
  Audio, Speech and Language}, vol. 117, 2013.

\bibitem{kim2019improved}
C.~Kim, M.~Shin, A.~Garg, and D.~Gowda, ``Improved vocal tract length
  perturbation for a state-of-the-art end-to-end speech recognition system,''
  \emph{Proc. Interspeech 2019}, pp. 739--743, 2019.

\bibitem{KSSdataset}
K.~Park, ``Kss dataset: Korean single speaker speech dataset,''
  https://kaggle.com/bryanpark/korean-single-speaker-speech-dataset, 2018.

\bibitem{griffin1984signal}
D.~Griffin and J.~Lim, ``Signal estimation from modified short-time fourier
  transform,'' \emph{Transactions on Acoustics, Speech, and Signal Processing},
  vol.~32, no.~2, pp. 236--243, 1984.

\bibitem{oord2016wavenet}
A.~v.~d. Oord, S.~Dieleman, H.~Zen, K.~Simonyan, O.~Vinyals, A.~Graves,
  N.~Kalchbrenner, A.~Senior, and K.~Kavukcuoglu, ``Wavenet: A generative model
  for raw audio,'' \emph{arXiv preprint arXiv:1609.03499}, 2016.

\bibitem{kingma2014adam}
D.~P. Kingma and J.~Ba, ``Adam: A method for stochastic optimization,''
  \emph{arXiv preprint arXiv:1412.6980}, 2014.

\bibitem{shi2016deconvolution}
W.~Shi, J.~Caballero, L.~Theis, F.~Huszar, A.~Aitken, C.~Ledig, and Z.~Wang,
  ``Is the deconvolution layer the same as a convolutional layer?'' \emph{arXiv
  preprint arXiv:1609.07009}, 2016.

\bibitem{keskin2019measuring}
G.~Keskin, T.~Lee, C.~Stephenson, and O.~H. Elibol, ``Measuring the
  effectiveness of voice conversion on speaker identification and automatic
  speech recognition systems,'' \emph{arXiv preprint arXiv:1905.12531}, 2019.

\end{thebibliography}


\end{document}